\documentclass[sigconf]{acmart}
\AtBeginDocument{%
  \providecommand\BibTeX{{%
    \normalfont B\kern-0.5em{\scshape i\kern-0.25em b}\kern-0.8em\TeX}}}

\copyrightyear{2023}
\acmYear{2023}
\setcopyright{acmlicensed}
\acmConference[CIKM '23] {Proceedings of the 32nd ACM International Conference on Information and Knowledge Management}{October 21--25, 2023}{Birmingham, United Kingdom.}
\acmBooktitle{Proceedings of the 32nd ACM International Conference on Information and Knowledge Management (CIKM '23), October 21--25, 2023, Birmingham, United Kingdom}
\acmPrice{15.00}
\acmISBN{979-8-4007-0124-5/23/10}
\acmDOI{10.1145/3583780.3614839} 

\pdfoutput=1
\usepackage{xcolor}
\definecolor{thedarkblue}{RGB}{0,0,120} 
\definecolor{mydarkblue}{rgb}{0,0.08,0.45} 
\definecolor{darkblue}{rgb}{0,0.08,180}
\colorlet{TufteRed}{red!80!black}
\definecolor{theblue}{RGB}{0,0,180}
\colorlet{thered}{TufteRed}

\usepackage[inline]{enumitem}
\usepackage{microtype}
\usepackage{algorithm} 
\usepackage[noend]{algpseudocode}

\usepackage{subfigure}
\usepackage{graphbox}
\usepackage{svg}
\usepackage{nicefrac}
\usepackage{csvsimple}

\usepackage{booktabs}
\usepackage{tabularx}

\usepackage{amsmath,amssymb,amsthm,bm}

\newcommand{\eat}[1]{\ignorespaces}
\usepackage{comment}

\newcommand{\journal}[1]{} 

\DeclareMathAlphabet{\mathbcal}{OMS}{cmsy}{b}{n}
\usepackage{mathrsfs}
\usepackage{graphicx}

\usepackage[utf8]{inputenc} 
\usepackage[T1]{fontenc}    
\usepackage{url}            
\usepackage{nicefrac}       
\usepackage{bbm}
\usepackage{listings}
\AtBeginDocument{
  \providecommand\BibTeX{{
    \normalfont B\kern-0.5em{\scshape i\kern-0.25em b}\kern-0.8em\TeX}}}

\usepackage{tikz}
\usepackage{verbatim}
\usetikzlibrary{arrows}
\usetikzlibrary{shapes,snakes}
\usetikzlibrary{decorations.pathmorphing} 
\usetikzlibrary{fit}					
\usetikzlibrary{backgrounds}	

\usepackage{ragged2e}
\usepackage{multirow}
\usepackage{microtype}
\usepackage{balance}
\usepackage{setspace}

\graphicspath{{./}{./graphics/}}
\newcolumntype{H}{>{\setbox0=\hbox\bgroup}c<{\egroup}@{}}

\newcolumntype{R}[1]{>{\RaggedLeft\arraybackslash}} 
\newcolumntype{L}[1]{>{\RaggedRight\arraybackslash}} 

\newcommand{\eg}{\emph{e.g.}}
\newcommand{\ie}{\emph{i.e.}}

\AtBeginEnvironment{pmatrix}{\setlength{\arraycolsep}{2pt}}

\DeclareMathOperator{\hugeE}{\mbox{\huge\raise-0.3ex\hbox{E}}}
\DeclareMathOperator{\p}{\mathbb{P}}
\DeclareMathOperator{\hugep}{\mbox{\huge\raise-0.3ex\hbox{$\p$}}}

\newcommand{\citedummy}[1]{\cite{#1}}
\newcommand{\dadsmin}{DAD-SMIN}
\newcommand{\f}{\mkern-2mu f\mkern-3mu}

\begin{document}

\title{Delivery Optimized Discovery in Behavioral User Segmentation under Budget Constraint}

\author{Harshita Chopra}
\affiliation{%
  \institution{Adobe Research, India}
  \country{}
  }
\email{harshitac@adobe.com}
\authornote{Equal Contribution.}

\author{Atanu R Sinha}
\authornotemark[1]
\affiliation{%
  \institution{Adobe Research, India}
  \country{}
  }
\email{atr@adobe.com}

\author{Sunav Choudhary}
\affiliation{%
  \institution{Adobe Research, India}
  \country{}
  }
\email{schoudha@adobe.com}

\author{Ryan A Rossi}
\affiliation{%
  \institution{Adobe Research, USA}
  \country{}
  }
\email{ryrossi@adobe.com}

\author{Paavan Kumar Indela}
\affiliation{%
  \institution{Adobe Research, India}
  \country{}
  }
\email{indela.paavankumar@gmail.com}
\authornote{The work was done while authors were in Adobe Research, India.}

\author{Veda Pranav	Parwatala}
\authornotemark[2]
\affiliation{%
  \institution{Adobe Research, India}
  \country{}
  }
\email{pvedapranav@gmail.com}

\author{Srinjayee Paul}
\authornotemark[2]
\affiliation{%
  \institution{Adobe Research, India}
  \country{}
  }
\email{srinjayeepaulsep@gmail.com}

\author{Aurghya Maiti}
\authornotemark[2]
\affiliation{%
  \institution{Adobe Research, India}
  \country{}
  }
\email{aurghya.kgp@gmail.com}

\renewcommand{\shortauthors}{Harshita Chopra, et al.}


\begin{abstract}
Users' behavioral footprints online enable firms to discover behavior-based user segments (or, segments) and deliver segment specific messages to users. Following the discovery of segments, delivery of messages to users through preferred media channels like Facebook and Google can be challenging, as only a portion of users in a behavior segment find match in a medium, and only a fraction of those matched actually see the message (exposure). Even high quality discovery becomes futile when delivery fails. Many sophisticated algorithms exist for discovering behavioral segments; however, these ignore the delivery component. The problem is compounded because (i) the discovery is performed on the behavior data space in firms' data (e.g., user clicks), while the delivery is predicated on the static data space (e.g., geo, age) as defined by media; and (ii) firms work under budget constraint. We introduce a stochastic optimization based algorithm for delivery optimized discovery of behavioral user segments and offer new metrics to address the joint optimization. We leverage optimization under a budget constraint for delivery combined with a learning-based component for discovery. Extensive experiments on a public dataset from Google and a proprietary dataset show the effectiveness of our approach by simultaneously improving delivery metrics, reducing budget spend and achieving strong predictive performance in discovery.

\end{abstract}



\maketitle

\section{Introduction}
The ubiquitous online user behavior data afford opportunities for behavioral user segmentation to online firms.
With ever more sophisticated algorithms, firms leverage the data of its own users to discover segments' propensities, behavioral tendencies, etc. to send messages, make predictions, offer recommendations, and improve users' experiences.
Behavioral user segments (hereafter, segments) are fundamental blocks for firms to target different segments with different messages and offerings (hereafter, messages)~\citedummy{gupta2014marketing}.
\textbf{Discovery} or formation of behavior segments, is only the first act and becomes ineffective unless messages can be actually delivered to the segments.
\textbf{Delivery} of messages to the segment is the second act.
Increasingly, delivery occurs through media channels (hereafter, media) such as Facebook, Google and others, as exemplified by media spend's strong growth approaching 300 billion USD~\citedummy{statista-ad-expense}.
Two uncertainties facing the firms thwart delivery: (1) only a portion of a behavior segment find \textbf{match} in a medium; and (2) only a fraction of those matched actually see the message, or has \textbf{exposure} to the message.
In media parlance~\citedummy{facebook}, when a message is sent to a segment, \textbf{Reach} occurs provided \textit{both} match and exposure are realized.
Yet, even advanced algorithms for discovery, unsupervised or supervised, ignore these uncertainties of reach
inherent in delivery~\citedummy{aljalbout2018clustering,chen2017purtreeclust,ezenkwu2015application,alkhayrat2020comparative,shin2019predictive,ghasemi2021survey,kameshwaran2014survey}.
Instead, those focus only on performance for discovery while tacitly assuming away these delivery roadblocks.
This is a problem and a research gap.

\begin{figure}[h!]
\centering
\includegraphics[width=0.8\linewidth]{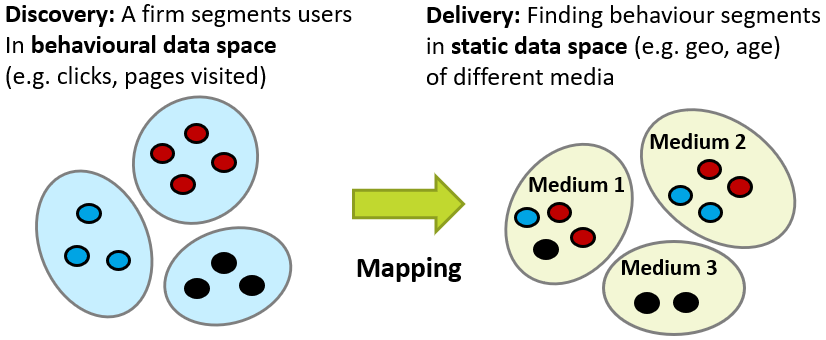}
\caption{Need for Delivery Optimized Segment Discovery}
\label{fig:deli_disc}
\end{figure}

Firms perennially face the problem of segmenting its users (customers) for effective targeting, and allocating segments thus formed to different media within a media budget~\cite{investopedia_ad}. Figure~\ref{fig:deli_disc} is a pictorial sketch of this problem.  
When the discovery is done independently of delivery consideration, suppose three behavior segments are formed.
Then, as delivery with static data is decided, a typical outcome is that each segment gets mapped to multiple media.
If the firm wants to deliver a message specifically aimed at the red-dots behavior segment, which is spread across two media, two problems arise: the budget can be exceeded, and the message is delivered to some users in blue-dot and black-dot segments, a potential waste.
To avoid these problems it would be prudent to account for delivery, during discovery of segments, by considering alignment with the three media.

The formal research problem is: 

1. How to improve match across media with the goal of improving reach for delivering communication to users?  

2. How to address (1) for endogenously formed audiences (as opposed to exogenously defined audiences as in the prior art) and multiple media?  

3. How to address (1) and (2) within a budget?  

4. How to also obtain high accuracy on predicting conversion? 

Achieving the \textit{dual goals of high conversion accuracy and reach maximization}, embedded in the research problem, is compounded by two other factors.
\begin{enumerate*}[label=(\alph*)]
	\item \label{intro:1} the discovery of behavior segments is performed on the firm's behavior data space (\eg~users' click, page visit, product viewed), all of which exist in users' data the firm possesses, but do not exist in the data of users the media possess.
	The media have users' static data (\eg~geo, age, interests) on which the delivery is predicated.
	\item	Firms work under tight budget constraints since media are costly.
\end{enumerate*}
Elaborating on \ref{intro:1}, users' visits to a firm's site or app generate a user behavior log of \textit{pageurls}, analyzing which users are clustered into behavior segments, in a supervised or unsupervised manner.
Behaviors are dynamic data.
For the same users, static data such as geo, age, and interests are available to the firm.
When a firm ports over a behavior segment to a medium for sending messages to the segment, the firm defines a set of static characteristics (\eg~\textit{country-us, age-35-40, interests-jazz}) that "best" represents the behavior segment.
Then this set, along with the message, are passed along to a medium, which does the delivery.
The medium relies on static data to find users with (\textit{country-us, age-35-40, interests-jazz}) from among the medium's vast user base.

A question may arise that the firm can bypass the behavior segmentation step and instead form segments in the static data space.
However, behavior data are much more predictive of a user's decision (\eg~whether to convert, whether to renew a subscription) than static data~\citedummy{gupta2014marketing}, reinforcing the premium put on behavior segments.
For effectiveness, it is necessary to maintain the primacy of behavior segmentation in the discovery phase and map to the static data for delivery.

Coming to (b), the firm can spend the media budget on various combinations of different behavior segments and different media.
Different behavior segments, defined by their sets of static characteristics, result in different proportions of match and exposure, and these proportions differ across different media.
Further, the cost of sending a message to a user varies by media.
Thus, a firm's spend depends on the composition of users in each behavior segment, those users' representation in static data (which determines match and the exposure proportions), and the cost per medium.
It follows that the discovery of behavior segments is intrinsically linked to the delivery and the spend; however, the prior art misses this linkage.

To address these limitations, we propose joint optimization of discovery and delivery for behavioral segmentation and optimize subject to the media spend budget constraint.
We address Reach, a common currency for media spend~\citedummy{facebook}, which maps to \textit{pay-per-click}. We focus on \textit{Direct}~\citedummy{amazon_ad} media spend and not on programmatic ad-bidding spend. 
For reach maximization subject to a budget, two different forms are modeled - one, based on mean squared error (MSE); and two, based on optimization under constraint~\citedummy{boyd2004convex}.
In each form we present multiple models; totaling five proposed models.
Moreover, since the space of behavior segments is different from that of static data, a separate network learns a mapping function \textit{Beh2Stat} such that a behavior segment can be expressed in terms of reach, through the match and exposure rates, to make stochastic assignments to the media.
Five performance metrics are used to span conversion prediction, spend and reach efficiency and effectiveness.  
We find that our proposed Delivery Aware Discovery (\textbf{DAD}) models produce predictive conversion accuracy, AUROC, comparable to models focused only on discovery (\textbf{DISC}), and yet reduces Spend and increases Reach. In addition, within our five proposed models, the Augmented Lagrangian stochastic optimization model has the best performance in closeness of spend to the budget, and better than that of the MSE based models.


\medskip\noindent\textbf{Summary of Main Contributions.}
Focusing on firm's \textit{direct} media spend~\citedummy{amazon_ad}, not programmatic spend, our key contributions are:
\begin{itemize}
    \item A new research problem of delivery aware discovery in behavioral user segmentation.
    \item A new model for joint optimization of discovery and delivery, under budget constraint.
    \item Recognizing the mapping from firm's behavioral data space to the media's static data space since behavioral segments are not realizable on media platforms.
    \item Performing stochastic optimization for the dual goals of (a) predictive segmentation and (b) optimizing reach.
    \item Introducing two new metrics for spend and reach efficacy.
\end{itemize}

\section{Related Work}
\label{sec:related-work}
Segmentation of users is a well-studied problem.
Clustering approaches are commonly used for segmentation, emanating from statistical cluster analysis in an influential 1963 paper~\citedummy{ward1963hierarchical}.
As online data have evolved, so have clustering algorithms ~\citedummy{aljalbout2018clustering,chen2017purtreeclust,ezenkwu2015application,alkhayrat2020comparative}.
Both unsupervised and supervised clustering methods dot the literature~\citedummy{shin2019predictive,ghasemi2021survey,kameshwaran2014survey}.
In particular, a recent paper~\citedummy{lee2020temporal} performs temporal predictive clustering on dynamic user data.
The outcome, conversion, is our prediction objective, for which segments are formed from event-level-behavior-sequence data (logs).
That is, our goal is not to cluster based on the behavior sequence per se, as papers~\citedummy{wang2016unsupervised} on unsupervised sequence clustering do.
A large literature of deep clustering~\citedummy{DBLP:journals/corr/XieGF15, DBLP:journals/corr/YangFSH16,DBLP:journals/corr/abs-1802-01059}, time series clustering~\citedummy{DBLP:journals/is/AghabozorgiST15,ratanamahatana2005novel}, and progression of diseases~\citedummy{DBLP:conf/ichi/RusanovPW16,DBLP:conf/ichi/LuongC17} do not address the research questions we tackle.
All these works address only the discovery of user segments, but none incorporates the delivery.
On the discovery side of the equation, the state of the art (SOTA) is the paper~\citedummy{lee2020temporal}.
Without any prior art for delivery aware discovery and our context of predictive clustering with users' dynamic, behavior data, we use~\citedummy{lee2020temporal} as the baseline SOTA model.

The problem of advertising (message) delivery under budget constraint occupies an early prominent role in search~\citedummy{karande2013optimizing,feldman2007budget}.
Optimization of message spend rate with budget is well attended~\citedummy{kumar2022optimal,agarwal2014budget}.
An area of research emphasis is budget pacing, whereby a given budget from a firm is dynamically allocated over a time horizon, based on anticipated traffic with specified characteristics, as exemplified by the empirical and theoretical contributions~\citedummy{xu2015smart,conitzer2022multiplicative,stier2018multiplicative,balseiro2017budget,balseiro2020dual}, to name a few.
This important body of research makes strong contributions toward managing delivery of messages to users within budget and addresses the context of ad bidding.
None of these works considers the discovery or formation of segments; they start with a given set of users.
We model discovery and delivery jointly.
Our work does not address the ad bidding based media spend; instead we address the direct media spend, whereby the cost of a message to a user is known and is not an outcome of bidding.
We also do not perform pacing of messages.


\begin{table}[t!]
\tiny
\caption{Statistics for the datasets used}
\centering
\resizebox{0.8\columnwidth}{!}{%
    \begin{tabular}{l|l|l}
    \toprule
     & Dataset I & Dataset II \\
    \hline
    \#Train Users & 1664 & 26292 \\
    \#Test Users & 416 & 6572 \\
    \hline
    \#Sessions per user: & {} & \\
    Min & 1 & 3 \\
    Max & 8 & 8 \\
    \hline
    \#Pages per session: & {} & \\
    Min & 2 & 5 \\
    Max & 40 & 40 \\
    \bottomrule
    \end{tabular}
}
\label{tab:data_stats}
\end{table}

\section{Data}

\subsection{Firm data}
The two datasets contain time stamped behavioral activity of users, for each user. Dataset I is proprietary, from a provider of SaaS services to individual users. Dataset II is public, from Google Analytics~\citedummy{google_data}.
For each dataset, we work with the site's pages (hereafter, page-urls), where the behaviors are captured in the format of page-urls. A user visits the site on multiple sessions, and in each session (visit) browses multiple pages. The logs show the time-stamped sequence of page-urls for each user as they click on pages, and are mapped to the user with an anonymized code.
Additionally, each dataset has a binary target label for conversion, corresponding to each session.
Statistics for the datasets are shown in Table~\ref{tab:data_stats}. These statistics are quite comparable to the industry standard \textit{average} number of pages per user, per session, ranging from 10 to 31~\citedummy{statista-pageviews}.

For the data at hand, behaviors are page-names in the form of page-urls, common for online browsing data. In other examples, such as campaign, behaviors can be open email, click email, unsubscribe, etc. Other target variables of interest to any firm can be used, including target variables with more than two classes.

Processed public data are available  \href{https://github.com/harshita219/Audience_Opt}{[here]}.

\subsection{Media data}
\label{sec:media_data}
Data of match rate and exposure rate are media dependent (\eg~different across Facebook, Google-YouTube, TikTok), and specific to the combination of static characteristics.
In our proprietary data, there are 5 static characteristics, (\textit{country, source, member-type, browser, os(operating system)}).
As an illustration, the 5-tuple, static characteristic set (\textit{country-us, source-bookmarked, member-frequent, browser-chrome, os-windows}) has a match rate 0.56 and exposure rate 0.47 for a medium $j$; while for another 5-tuple (\textit{country-uk, source-bookmarked, member-infrequent, browser-edge, os-iOS}) those values are 0.39 and 0.61 for the same medium $j$.
However, in another medium $j\prime$,  the former 5-tuple (\textit{country-us, source-bookmarked, member-frequent, browser-chrome, os-windows}) has values 0.45 and 0.69.
The match and exposure rates vary across the combinatorial set of static characteristics, for each medium.
These data are available from a medium to the firm which advertises on the medium through reporting APIs but are \textit{not publicly} disclosed.
To overcome this, we move to generate a match rate and exposure rate from a joint distribution over the support of the combinatorial set of static characteristics, which implies that for every combination (tuple) of the 5 static features, a match rate and an exposure rate, lying between $0.25$ and $0.75$ are drawn, for a medium.
This is repeated for each medium.
The match and exposure rates are held fixed for all the models.
The cardinality of the set of 5-tuple static characteristics in our data is $1,296$. We thus have a table of pairs of ({match rate, exposure rate}) for each $1,296$ five tuples and three media. 

From the Google data, we use 6 static features, each categorized as follows: (\textit{medium [organic, referral/cpc/cpm, others]; deviceCategory [desktop, mobile/tablet]; 
operatingSystem [Macintosh/iOS, Windows, others];  
categorized-geo [North America, Asia, Europe, others];  
browser [Chrome, Safari and others]; source [direct, Google, others]}). Match and exposure rates for the Google data are generated in the manner described in the above paragraph. These static characteristics and their categories yield a table of $432$ six tuples and three media, giving us pairs of (match rate, exposure rate) for each of $432$ six tuples, for each of three media.

\begin{figure*}[h!]
\centering
\includegraphics[width=0.95\linewidth]{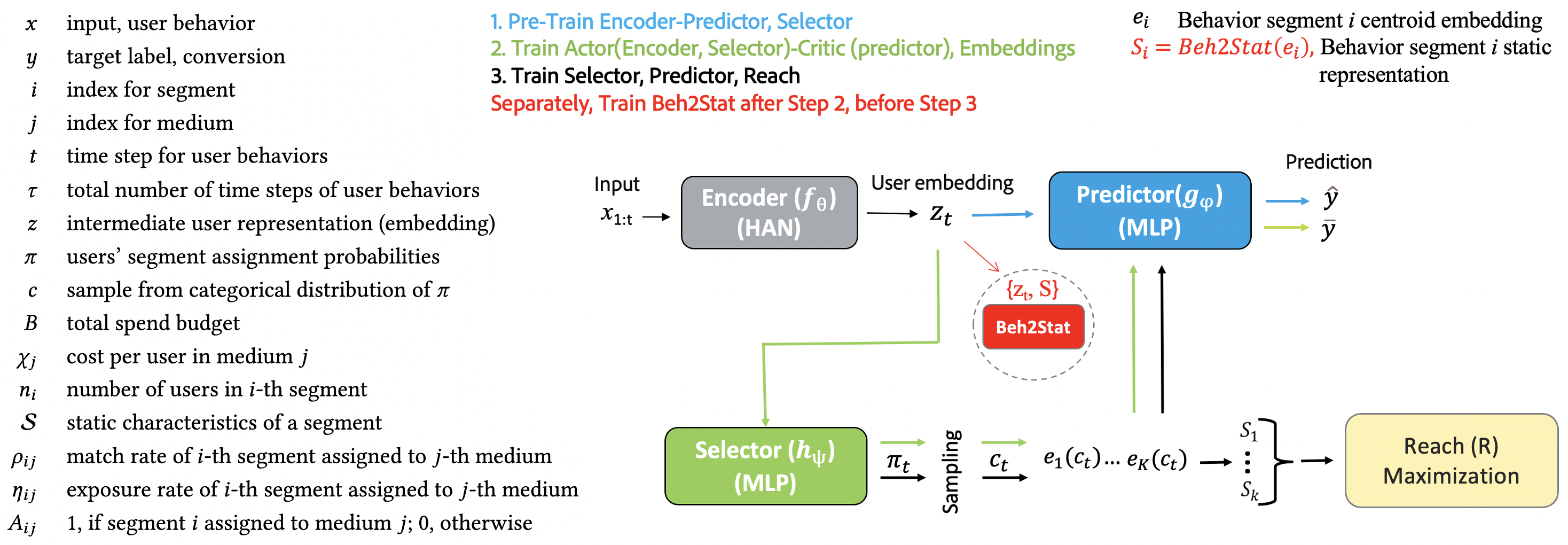}
\caption{Overview of approach.}
\label{fig:overview}
\end{figure*}

\section{Joint Discovery-Delivery Model Architecture}
To achieve the dual goals of (i) conversion prediction and (ii) reach maximization subject to a budget, we introduce a two-part network. Figure~\ref{fig:overview} summarizes the notation and depicts the model architecture. In the Figure, green colored network is for goal (i), and black for goal (ii). Additionally, as a necessity to map behavioral data of users to static characteristics required by media, we introduce a mapping function \textit{Beh2Stat}. The match rate and exposure rate vary by static characteristics and medium. The cost varies by medium. The \textit{Beh2Stat} function’s mapping to the static space, allows computation of cost of reach, which is then taken to the budget constraint. Training the whole network in three steps as shown in Figure 2 achieves the joint optimization. The SOTA in discovery~\citedummy{lee2020temporal} is preserved by green network. 

\subsection{HAN Encoder}
\textbf{Encoder} $f_\theta$: $\prod_{i=1}^{t} \mathcal{X} \to \mathcal{H}$.
The encoder takes user behavior till time $t$ as input represented by $\mathbf{x}_{1:t} = \{x_1, x_2, \dots, x_t\}$ for $\mathbf{x}_t {\in} \mathcal{X}$, and learns to produce an intermediate user-level representation $\mathbf{z}_t {\in} \mathcal{H}$, by predicting each user's target label, as in, $\hat{y}_t$.
Note $\mathbf{z}_t$, a hidden vector in the latent space $\mathcal{H}$, is a representation that embodies the latent tendency of a user, and is used for the clustering task.
The encoder used here is a Hierarchical Attention Network (HAN)~\citedummy{yang-etal-2016-hierarchical}.
We choose HAN since it better encapsulates the two-level sequence of user behaviors - one, multiple pages browsed in each session and two, multiple sessions of each user.
The first level encodes activities within each session to a session-level vector and the second level encodes the session-vectors to a user-level vector.

\subsection{Predictor, Selector, Embedding Dictionary}
\textbf{Predictor} $g_\phi$: $\mathcal{H} \to [0,1]$ is a fully connected network that takes an embedding $z$ as input and predicts the target, i.e., conversion probability $y$.	\\
\textbf{Selector} $h_\psi$: $\mathcal{H} \to \Delta^{K-1}$ is a fully connected network that takes a user embedding $z$ as input and computes a distribution $\pi$ where $\pi(k)$ is the probability of the embedding $z$ being assigned to the $k$-th cluster.	\\
\textbf{Embedding Dictionary} $\mathcal{E}$: This is a dictionary of the centroids of $K$ clusters.
Given a sampled cluster assignment $c_t$, it outputs the centroid embedding $\mathbf{e}(c_t) \in \mathcal{H}$.

\subsection{Beh2Stat}
$\textbf{Beh2Stat}$. $b_\omega$: $\mathcal{H} \to \mathcal{S}$ is a fully connected network that learns the function mapping user behavioral embedding $z_t$ to the user static characteristics vector $S \in \mathcal{S}$.
Note that, by definition, \textit{static} characteristics of a user do not depend upon $t$.
Once trained, this function projects the $i$-th behavioral segment-specific centroid embedding $\mathbf{e}(c_t) \in \mathcal{H}$ to the $i$-th segment-specific static characteristics $S_i \in \mathcal{S}$.
The projection is necessary to assign match rate $\rho_{ij}$ and exposure $\eta_{ij}$ to the $i$-th segment, since for any medium $j$, $\rho_{ij}$ and $\eta_{ij}$ are defined in terms of static characteristics $S_i$, but not in terms of behavioral embedding.
The $\rho_{ij}$ and $\eta_{ij}$ are used in the reach computation, defined later.
Specifically, to preserve an important property, that the behavioral segment contains users with a variety of static characteristics, we project a probability distribution $p$ over the support of $\mathcal{S}$ and use that probability distribution to compute expected reach.
This affords stochastic optimization of the objective, reach.
The loss associated with Beh2Stat is given by $\mathcal{L_B} (\omega) = - \sum_{} S \log p(S)$.


\section{Objective Function and Constraints}
\label{sec:budget_opt}

\subsection{Behavioral Segmentation Objective}
\label{beh_segmentation}
The input $\mathbf{x}_{1:t}$ are user behaviors present in the sequence of page urls clicked.
The output $y_t$ of a user denotes whether a conversion was observed or not.
The objective of behavioral segmentation is to cluster users into segments based on users' embeddings $\mathbf{z}_t$ such that segment-wise average prediction of conversion performs well.
The following describes the training.

\subsection{Reach Maximization Objective}
We assume that each segment is assigned to only one medium (\ie~one media channel), but multiple segments can be communicated through the same medium.
During training, an MLP $v_\delta(\mathcal{S})$ learns to map static features to mediums.
For $i$-th segment's  static characteristics $S$, given a medium $j$, match rate and exposure rate coming from a table of match and exposure rates, as stated in Section~\ref{sec:media_data}.
The \textit{reach} objective is denoted by $\mathcal{L}_{R}$ and is calculated as follows.

The loss, $\mathcal{L}_{4}$, for the joint optimization of discovery and delivery is,
\begin{equation}
    \mathcal{L}_4(\theta, \phi, \psi, \omega, \delta) = \mathcal{L}_A(\theta, \psi, \phi) + \mathcal{L}_C(\phi) + \mathcal{L}_{R}(\delta)
\end{equation}
where, $\mathcal{L}_{R}$ is the loss for the reach maximization.
Note that $\mathcal{L}_{R}$ changes with the specific formulation of the optimization's objective function. $\mathcal{L}_A$ denotes Actor loss and $\mathcal{L}_C$ denotes Critic loss, both of which are defined in Section~\ref{Sec:Training}.


Since each segment is activated on only one channel, the channel should be selected to maximize the reach of that segment.
Hence, $i$-th segment's reach $R_i = \max_{j}(A_{ij}\rho_{ij} \eta_{ij}$).
Expected total reach $R$ is the sum of the reaches of individual segments and is given by
\begin{equation}
    R = \sum_{i} R_i = \sum_{i} \max_{j} (A_{ij}\rho_{ij} \eta_{ij}) n_i
    \label{eqn:reach}
\end{equation}

\subsubsection{Mean Squared Error (MSE)}
\label{sec:MSE}
The first form of reach maximization is based on MSE.
The total budget is split among the different media proportionate to the number of people it contains, which is the sum of the number of people in all segments assigned to that media.
Depending upon match rate $\rho_{ij}$, and exposure rate $\eta_{ij}$ of the static characteristic tuple of the segment to which the user belongs, the spend for reach differs across users in different segments and across different media.
Note that ($\rho_{ij}, \eta_{ij}$) vary by static characteristic tuples and by media.
Segments formed through the Selector yield the size $n_i$, for the $i$-th segment, where $i=1,...,K$, and $K$ is the number of segments.
The Selector yields the group of users in segment $i$, whose latent, segment-centroid embedding $\mathcal{E}_i$ is passed to the \textit{Beh2Stat}, which outputs $i$-th segment's static characteristics tuple.
Given this tuple, corresponding ($\rho_{ij}, \eta_{ij}$) for each medium $j$ are read from a table for segment $i$.
For MSE, per individual user in segment $i$, medium $j$, the expected reach =  $\rho_{ij} \eta_{ij}$.
Per individual user assigned to medium $j$, the reach goal =  $\nicefrac{B}{N \chi_j}$.
The intuition is that with $N$ users, and cost per user reached $\chi_j$, the budget $B$ is divided into a reach goal per user.
Two variations of MSE are modeled as loss functions and described below.

\textbf{Cluster Specific Reach MSE}:
A loss is defined per cluster and the back propagation is per cluster.
The loss for $i$-th cluster is,
\begin{equation}
    \mathcal{L}_{R} = \sum_{j}A_{ij}\bigg(\rho_{ij} \eta_{ij} - \nicefrac{B}{N \chi_j}\bigg)^{2}
    \label{eqn:cluster_specific_error}
\end{equation}

\textbf{Cluster Agnostic Reach MSE}:
Here the loss is averaged across all clusters, and the back propagation is across all clusters.
The MSE is,
\begin{equation}
    \mathcal{L}_{R} = \frac{\sum_{i}\sum_{j}A_{ij}n_i * \bigg(\rho_{s_{ij}} \eta_{s_{ij}} - \nicefrac{B}{N \chi_j} \bigg)^{2}}{\sum_{i} n_i}
    \label{eqn:cluster_agnostic_error}
\end{equation}

\subsection{Budget Constraint}
\label{impression_budget}
Let $j' \triangleq \arg\max_{j} (A_{ij} \rho_{ij} \eta_{ij})$ be the channel which maximizes $R_i$ for any given segment $i$.
The budget constraint for total reach is
\begin{equation}
    T_R = B - \sum_{i} (A_{ij'} \rho_{ij'} \eta_{ij'}) \chi_{j'} n_i \geq 0
    \label{eqn:constraint}
\end{equation}
where the second term in equation~\eqref{eqn:constraint} is the \textbf{\textbf{Spend}}.

\section{Training} \label{Sec:Training}
Pseudo-code of the model is given in Algorithm~\ref{algo_main}, and the joint optimization in Algorithm~\ref{algo_joint_opt}.
Top left of Figure~\ref{fig:overview} mentions the ordered sequence of steps needed for initialization and training.

Pre-training and Initialization consists of doing the following (Algorithm~\ref{Pre-training}).
\begin{enumerate}
	\item	Pre-train the Encoder and the Predictor using the loss
	\begin{equation}
	    \mathcal{L}_1 (\theta, \phi) = \mathbb{E}_{\mathbf{x},y \sim p_{XY}} \bigg[-\sum_{t \in \mathcal{T}} l_1(y_t, \hat{y}_t)\bigg]
	    \label{EP_loss}
	\end{equation}
	where $\hat{y}_t = g_\phi(f_\theta(\mathbf{x}_t))$ is the predicted conversion probability of a user and $l_1(y_t, \hat{y}_t) = -\sum\limits_{c\in \{0,1\}} y_{t}^{c} \log(\hat{y}_{t}^{c})$.
	\item	Initialize the cluster embeddings using K-means on the representations $z^{(n)}_t$ for all $n$ users and for all $t$ that are obtained after pre-training the encoder.
	\item	Pre-train the Selector on all $z^{(n)}_t$ and corresponding cluster assignments obtained from K-means.
\end{enumerate}

Subsequently, we train as follows.
Lines 1-8 in Algorithm~\ref{algo_main} use an alternating minimization approach to alternate between training an Actor-Critic network and updating the embedding dictionary.
Here the Actor is the (Encoder, Selector) pair of networks and the Critic is the Predictor network.
Lines 9-13 in Algorithm~\ref{algo_main} train the Beh2Stat network.
Finally, lines 14-19 use alternating minimization to alternate between maximizing the reach and updating the embedding dictionary.

The actor's loss is $\mathcal{L}_A(\theta, \phi, \psi) = \mathcal{L}_1 (\theta, \phi, \psi) + \alpha \mathcal{L}_2 (\theta, \phi)$ which combines two losses with $\alpha$ as the hyperparameter.
The loss term $\mathcal{L}_2 (\theta, \psi)$ promotes sparse cluster assignment such that each user belonging to only one cluster with high probability.
It is given by
\begin{equation}
	\mathcal{L}_2 (\theta, \psi) = \mathbb{E}_{\mathbf{x} \sim p_X} \bigg[-\sum_{t \in T}\sum_{k \in K}\pi_t(k) \log \pi_t(k) \bigg]
	\label{pis_cross_entropy_loss}
\end{equation}
The loss term $\mathcal{L}_1 (\theta, \phi, \psi)$ promotes the prediction of cluster level outcomes $\Bar{y}_t$ from the cluster centroid.
It is the partial function obtained by fixing the embedding $\mathcal{E}$ is the loss expression $\mathcal{L}_1 (\theta, \phi, \psi, \mathcal{E})$ given by
\begin{equation}
 \mathcal{L}_1 (\theta, \phi, \psi, \mathcal{E}) = \mathbb{E}_{\mathbf{x},y \sim p_{XY}} \bigg[ \sum_{t \in T} \mathbb{E}_{c_t \sim Cat(\pi_t)} \big[l_1(y_t, \Bar{y}_t) \big] \bigg]
 \label{critic_loss}
\end{equation}


The critic's loss is set to $\mathcal{L}_C(\phi) = \mathcal{L}_1(\theta, \phi, \psi)$.
Further, to promote well-separated cluster centroids in the embedding dictionary representation, the loss $\mathcal{L}_E (\mathcal{E}) = \mathcal{L}_1 (\mathcal{E}) + \beta \mathcal{L}_3 (\mathcal{E})$ is used to update the embedding dictionary, where
\begin{equation}
    \mathcal{L}_3 (\mathcal{E}) = -\sum_{k \neq k'} l_1 (g_\phi(\mathbf{e}(k), g_\phi(\mathbf{e}(k'))
    \label{embedding_loss}
\end{equation}



For the reach maximization subject to budget constraints, we use the known techniques of Barrier method and Augmented Lagrangian from constrained deterministic optimization~\citedummy{boyd2004convex} to convert it to an equivalent unconstrained objective for Algorithm~\ref{algo_joint_opt}.
These methods are fairly well understood for convex optimization.
However, neural network training is a non-convex problem and in this setting, the methods employing Barriers or Augmented Lagrangians are not as well understood, despite being intuitive.
The three formulations that we implement are
\begin{enumerate}
	\item	\label{sec:opt_lagrangian}
	\textit{Slack Minimization} with loss
	\begin{equation}
	    \mathcal{L}_{R} = \frac{1}{\ln(R)} + \frac{max(T_R,0)}{w}	\label{obj:constr_opt}
	\end{equation}
	The dual variable update rule is $w \gets \mu * w$ with initialization $\mu \gets 0.3$.

	\item	\label{sec:opt_barrier_method}
	\textit{Barrier Method} with loss
	\begin{equation}
	    \mathcal{L}_{R} = \frac{1}{\ln(R)} - \frac{\log( -T_R)}{w}	\label{obj:barrier}
	\end{equation}
	and the same dual update rule and initialization as above.

	\item	\label{sec:opt_augmented_lagrangian}
	\textit{Augmented Lagrangian Method} with loss
	\begin{equation}
	    \mathcal{L}_{R} = \frac{1}{\ln (R)} - \frac{\lambda T_R}{B} + \frac{\mu}{2}\ {\bigg(\frac{T_R}{B}\bigg)}^2
	    \label{obj:alm}
	\end{equation}
	The update rule is $\lambda_{k} \gets max(\lambda_{k-1} +\mu*\frac{T_{R, k-1}}{B},0)$, where $k$ is iteration.
	Note that $\lambda \geq 0$ and we use initialization $\mu \gets 0.1$, $\lambda \gets 0.1$.
\end{enumerate}

\begin{algorithm*}[!ht]
\caption{Pre-training} \label{Pre-training}

\begin{algorithmic}
\State \textbf{Input:} Dataset $\mathcal{D} = \{(\mathbf{x}_t^n, y_t^n)_{t=1}^{\tau^n}\}_{n=1}^N$, number of clusters $K$, learning rate $\eta$, mini-batch size $n_{mb}$
\State \textbf{Output:} Model parameters $\{\theta, \psi, \phi\}$, initialized by Glorot-Uniform, embedding dictionary $\mathcal{E}$.
\end{algorithmic}

\begin{algorithmic}[1]
\Statex \textit{Pre-train Encoder-Predictor} 
\Repeat
    \State Sample mini-batch of $n_{mb}$ samples
    \For {n = 1 to $n_{mb}$}
        \State Calculate $\hat{y}_t^n \leftarrow g_{\phi}(f_{\theta}(\mathbf{x}_{1:t}^n))$
    \EndFor
    \State $\theta \leftarrow \theta - \eta \frac{1}{n_{mb}} \sum_{n=1}^{n_{mb}} \sum_{t=1}^{T^n} \nabla_\theta l_1(y_t^n, \hat{y}_t^n)$, $~~~~\phi \leftarrow \phi - \eta \frac{1}{n_{mb}} \sum_{n=1}^{n_{mb}} \nabla_\phi l_1(y_t^n, \hat{y}_t^n)$
\Until{convergence}
\Statex Calculate embeddings dictionary $\mathcal{E}$ and initial cluster assignments $c^n_t$ 
\State \hspace{2cm} $\mathcal{E}, \{\{c_t^n\}_{t=1}^{\tau^n}\}_{n=1}^N \leftarrow \texttt{K-Means}(\{\{z_t^n\}_{t=1}^{n_{mb}}\}_{n=1}^N)$, \hspace{0.5cm} where $z_t^n = f_\theta (\mathbf{x}_{1:t})$
\newline\textit{Pre-train the Selector}
\Repeat
    \State Sample a minimbatch $n_{mb}$ of data samples
    \For {$n = 1 \ldots n_{mb}$}
        \State Calculate cluster assignment probability $\mathbf{\pi}_t^n \leftarrow h_\psi (f_\theta (\mathbf{x}_{1:t})$
    \EndFor
    \State Update selector parameters, $\psi \leftarrow \psi + \eta \frac{1}{n_{mb}} \sum_{n=1}^{n_{mb}} \sum_{t=1}^{\tau^n} \sum_{k=1}^K c_t^n(k) \log \pi_t^n(k)$
\Until{convergence}
\end{algorithmic}

\end{algorithm*}

\begin{algorithm}
    \begin{algorithmic}[1]
    \caption{Pseudo-code}
    \label{algo_main}
    \Statex \textit{Pre-train Encoder-Predictor}
    \Statex \textit{Initialize Cluster Centroid Embeddings}
    \Statex \textit{Pre-train Selector}  \Comment{Run Algorithm~\ref{Pre-training}}

    \Repeat
        \State Sample mini-batch of $n_{mb}$ samples
        \For {n = 1 to $n_{mb}$}
            \State Train Actor $\leftarrow \min \mathcal{L}_A(\theta, \phi, \psi)$
            \State Train Critic $\leftarrow \min \mathcal{L}_C(\phi)$
        \EndFor
        \For {n = 1 to $n_{mb}$}
            \State Update embedding dictionary $\leftarrow \min \mathcal{L}_E(\mathcal{E})$
        \EndFor
    \Until{convergence}
    \Repeat
        \State Sample mini-batch of $n_{mb}$ samples
        \For {n = 1 to $n_{mb}$}
            \State Train Beh2Stat $\leftarrow \min \mathcal{L}_B(\omega)$
        \EndFor
    \Until{convergence}
    \Repeat
        \State Sample mini-batch of $n_{mb}$ samples
        \For {n = 1 to $n_{mb}$}
            \State Train Joint\_Opt $\leftarrow \min \mathcal{L}_4(\theta, \phi, \psi, \omega, \delta)$
            \State Update embedding dictionary $\leftarrow \min \mathcal{L}_E(\mathcal{E})$
        \EndFor
    \Until{convergence}
    \end{algorithmic}
\end{algorithm}

\begin{algorithm}
    \begin{algorithmic}[1]
    \caption{Joint\_Opt training}
    \label{algo_joint_opt}
        \For {each mini-batch in $\{{(x^n_t,y^n_t)}^{\tau^{n}}_{t=1}\}^{n_{mb}}_{n=1} \sim \mathcal{D}$}
                \State ${z}_t^n \leftarrow f_{\theta}(\mathbf{x}_{1:t}^n)$
                \State ${\pi}_t^n \leftarrow  h_\psi({z}_t^n)$
                \State sample cluster assignment ${c}_t^n \sim  Cat({\pi}_t^n))$
                \State centroid embedding $\Bar{z}_t^n \coloneqq \mathcal{E}({c}_t^n)$
                \State $\Bar{y}_t^n \leftarrow  g_\phi(\Bar{z}_t^n)$
                \State $\mathcal{S} \leftarrow  b_\omega(\Bar{z}_t^n)$
                \State $A_{ij} \leftarrow$ $v_\delta(\mathcal{S})$
                \State Compute reach, $R$ using \eqref{eqn:reach}
                \State Compute constraint, $T$ using \eqref{eqn:constraint}
            \State Update $h_\psi$, $g_\phi$ and $v_\delta$, minimize $\mathcal{L}_4(\theta, \phi, \psi, \omega, \delta)$
        \EndFor
    \end{algorithmic}
\end{algorithm}

\section{Implementation Details}
\begin{figure}[h!]
\centering
\includegraphics[width=0.8\linewidth]{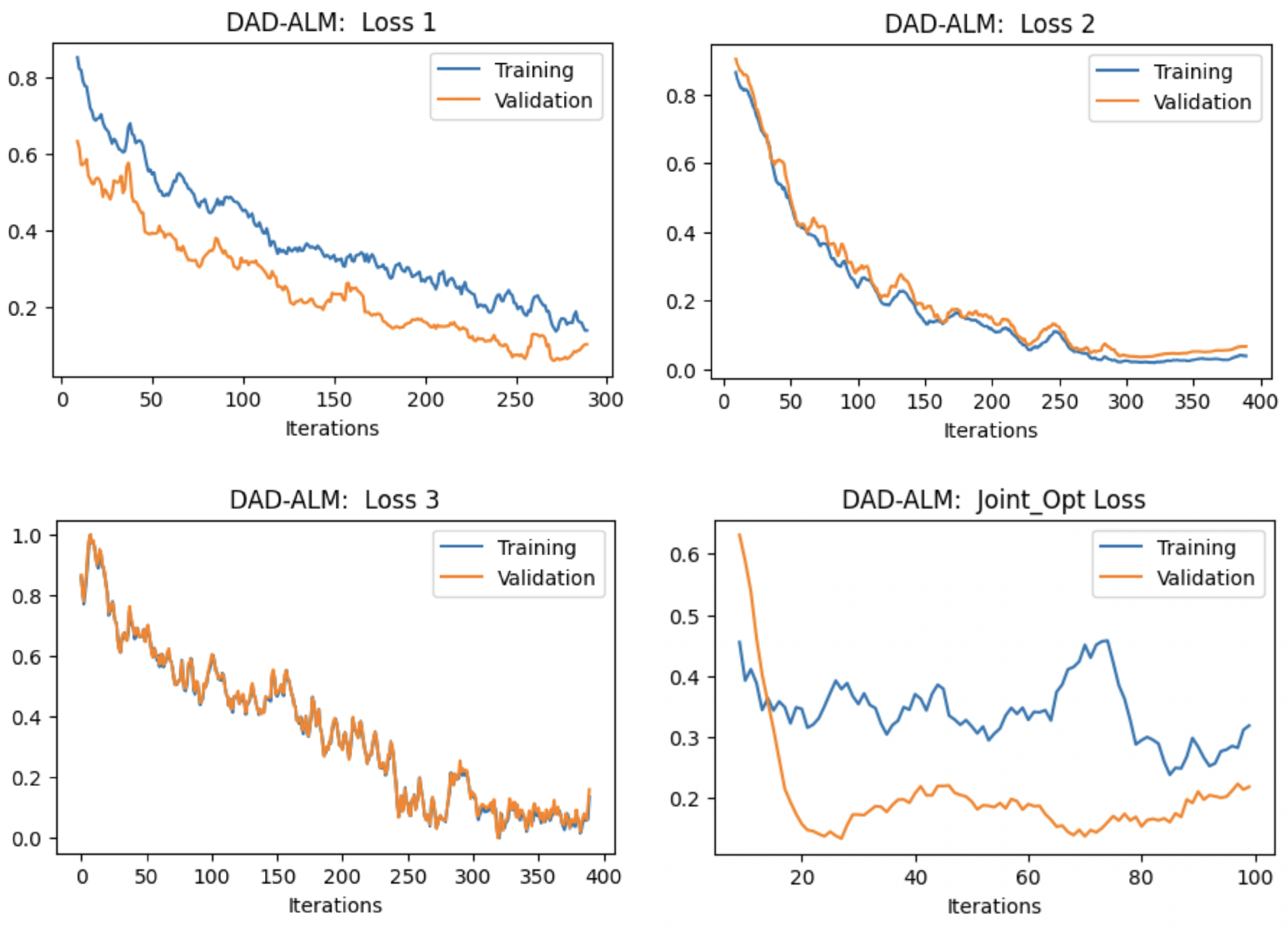}
\caption{Convergence plots for Losses $\mathcal{L}_{1}$, $\mathcal{L}_{2}$, $\mathcal{L}_{3}$ and $\mathcal{L}_{4}$. DAD-ALM, Dataset II. Training - Blue, Validation - Orange.}
\label{fig:google_imp_alm_plots}
\end{figure}

Our experimental setup including network parameters and hyperparameters used are as follows.
The encoder is a HAN, with the dimension of the hidden layer being $50$.
The predictor is a Multi-Layered Perceptron (MLP), with input $z_t$ (of dimension $50$), two hidden layer with $50$ perceptrons and an output layer of size $1$.
A dropout rate of $0.3$ is used after the hidden layer.
Hidden layers have ReLU activation and the output layer has sigmoid activation.
Selector is also an MLP with input $z_t$, followed by hidden layer with $50$ perceptrons, which uses ReLU activation and a dropout $0.3$.
The output layer is of size $K$, with softmax activation.
The Encoder-Predictor, Selector, Actor and Critic are trained using Adam Optimizer with a learning rate of $0.001$ on the aforementioned loss functions.
The network weights and biases are initialized using `glorot uniform'.
Batch size used is 128.
Initialization iterations for pretraining the Encoder-Predictor is 1000 and that for pretraining the selector is 5000.
Number of iterations for training actor, critic and embeddings is 1000, with an early stopping after 100 epochs, based on minimum value of $\mathcal{L}_1$, $\mathcal{L}_2$ and $\mathcal{L}_3$ obtained on validation data within previous 15 iterations.
Beh2Stat is an MLP with input $z_t$ (of dimension $50$), four hidden layer with $500$ perceptrons and the output layer uses Softmax activation to yield probabilities over the static features.
It is trained for the same number of iterations as actor-critic, using Adam Optimizer with a learning rate of $0.005$.
The joint optimization model (Algorithm \ref{algo_joint_opt}) is trained to update the parameters of Selector, Predictor and MLP $v_\delta(\mathcal{S})$, with the input as static features and output as mediums.
It is trained for $2000$ iterations, with an early stopping after 100 epochs, based on minimum value of $\mathcal{L}_4$ obtained on validation data within previous 15 iterations. Depicting training and validation convergence plots for Augmented Lagrangian Method on Dataset II, Figure~\ref{fig:google_imp_alm_plots} shows good convergence for the four losses.

\section{Experiments}
\label{sec:exp}
We design experiments around joint optimization of two goals: (i) strong predictive conversion performance for behavior segments; and (ii) achieve high reach relative to spend, subject to budget constraint. In direct media spend, one can think of reach as corresponding to pay-per-click, used by businesses to pay for messages they send to the targeted segments. 
Our experiments test these goals and are described next.

\subsection{Experimental Setup}
\label{sec:exp-setup}
Two SOTA baselines (\textbf{DISC}) plus five proposed \textbf{D}elivery-\textbf{A}ware-\textbf{D}iscovery (\textbf{DAD}) models are now presented.

\subsubsection{Baselines}
\begin{itemize}
    \item \textbf{DISC-UC}: The first baseline, based on behavioral segmentation objective in Section~\ref{beh_segmentation}, is the \textbf{disc}overy focused SOTA baseline, not delivery-optimized, since delivery-optimized-discovery is not available in the prior art. Here, the spend is unconstrained (\textbf{UC}) by the budget.
    \item \textbf{DISC-BC}: The second baseline, is the \textbf{disc}overy focused SOTA baseline, not delivery-optimized, but the spend is budget constrained (\textbf{BC}). We train Step 1 and Step 2 of the network to obtain $K$ segments, and skip Step 3 reach maximization. We compute reach for each segment, for each medium, and assign each segment to a medium giving the highest reach value for it. This process is repeated 
    until no budget is left over. 
\end{itemize}

\subsubsection{Proposed Models}
\begin{itemize}
    \item \textbf{DAD-CSSE}:  Cluster Specific MSE, equation \eqref{eqn:cluster_specific_error}
    \item \textbf{DAD-CASE}: Cluster Agnostic MSE, equation  \eqref{eqn:cluster_agnostic_error}
    \item \textbf{\dadsmin}: Slack Minimization, equation  \eqref{obj:constr_opt}
    \item \textbf{DAD-BARR}: Barrier, equation  \eqref{obj:barrier}
    \item \textbf{DAD-ALM}:  Augmented Lagrangian, equation  \eqref{obj:alm}
\end{itemize}

\subsubsection{Performance Metrics}
We evaluate against the ground truth of outcome, conversion, and spend efficiency to increase reach. We introduce two new metrics which combine conversion accuracy, spend, and budget to give new insights about efficiency and effectiveness of discovery and delivery.

\textbf{AUROC.}
Consistent with the first goal of strong predictive conversion for behavior segments, the performance metric is $AUROC_{\overline{y}}$, for which \textit{higher values are better}.

\textbf{Spend per Unit Reach.}
The spend metric comparable across models including baselines is the spend per unit reach. Since delivery-optimized-delivery is not available in the two DISC baselines an overall spend metric is unfair to baselines. However, spend per unit reach can be used since both reach and spend are affected in a comparable way for each model. Here, \textit{lower values are better}.

\textbf{Within \% Budget.}
This metric expresses the optimal spend relative to the budget and applies to all five proposed DAD models. Since the DISC-UC baseline's delivery has no budget constraint, the metric is undefined for DISC-UC. The DISC-BC baseline is budget constrained and this metric applies. Here, a \textit{tighter interval around zero is better} since it indicates closeness to the budget.

\textbf{Effective Spend.}
In a new contribution, this metric, capturing 'effectiveness of spend,' combines the two metrics $AUROC_{\overline{y}}$ and $Spend\ per\ Unit\ Reach$, used respectively for the two goals of high predictive accuracy and low spend per unit reach (Section~\ref{sec:exp}), into a composite scalar metric. By combining two metrics, a scalar metric represents the combined goal. We define \textit{Effective Spend} = \textit{Spend per Unit Reach} / $AUROC_{\overline{y}}$. For the numerator, lower is better, while for the denominator, higher is better. Put together, for Effective Spend, \textit{lower values are better}.  

\textbf{Reach Efficiency-Effectiveness, or, Reach Effc-Effe.}
In contributing another new metric, this is the second composite, scalar metric which captures both the 'efficiency' and 'effectiveness' of reach. Note that we seek to maximize reach, subject to a budget. We define, \textit{Reach Effc-Effe} = (\textit{Reach} / \textit{Spend as Proportion of Budget}) * ($AUROC_{\overline{y}}$). The term in the first parenthesis represents reach-efficiency, making reach-efficiency increase (decrease) when the denominator is less (more) than 1. The term in the second parenthesis stands for accuracy of prediction, and thus combined with the first term, provide a succinct representation of both the efficiency of budget spend to maximize reach, and the effectiveness in achieving the other goal of high performance in predictive accuracy. The DISC-UC baseline's delivery has no budget constraint; this metric is undefined for DISC-UC. Here, \textit{higher values are better}.

\subsection{Results, Dataset I}
\label{sec:reach_data1}
All results are shown with \textbf{Mean +/- Stderr} based on 8 seeds. Caption of Table~\ref{proprietary_data_performance_reach} highlights the significantly improved performance of proposed \texttt{DAD} models over SOTA baselines \texttt{DISC}s. Note that two metrics are undefined for DISC-UC since it is unconstrained. Overall, DAD-SMIN and DAD-ALM, each with good performance in 3 metrics, including two important composite metrics \textit{Effective Spend} and \textit{Reach-Effc-Effe}, stand out. Thus, decreased spend for maximizing reach is achievable with no decrease in conversion performance.

\begin{table}[hb]
\centering
\caption{
Results. Dataset I. K=5. In $AUROC_{\overline{y}}$ (higher is better), five proposed \texttt{DAD} models perform identical to the two \texttt{DISC} baselines. In $Within\ \%\ Budget$ (closer to 0 is better), baseline \texttt{DISC-BC} outperforms \texttt{DAD} models. In $Spend\ per\ unit\ Reach$ (lower is better), \texttt{DAD-SMIN}, \texttt{DAD-ALM} outperforms \texttt{DISC} baselines. In $E{\f}{\f}ective\ Spend$ (lower is better), \texttt{DAD-SMIN}, \texttt{DAD-CSSE}, \texttt{DAD-CASE}, \texttt{DAD-ALM} outperform \texttt{DISC} baselines. In $Reach\ E{\f}{\f}c$-$E{\f}{\f}e$ (higher is better), \texttt{DAD-SMIN}, \texttt{DAD-CSSE}, \texttt{DAD-CASE}, \texttt{DAD-ALM} outperform \texttt{DISC-BC} baseline. Overall, the \texttt{DAD} models show strong and significantly improved performance (shown in bold) over \texttt{DISC} baselines. See Section~\ref{sec:reach_data1} for discussion.}
\resizebox{\columnwidth}{!}{%
\begin{tabular}{l|c|c|c|c|c}
\toprule
Model & AUROC  & Within & Spend per & Effective & Reach \\
    {} & $\overline{y}$ & \% Budget & unit Reach & Spend & Effc-Effe \\
\midrule
DISC-UC & $0.873$ & -- & $66.993$ & $76.751$ & -- \\
& $\pm0.002$ & -- & $\pm1.42$ & $\pm1.722$ & -- \\
DISC-BC & $0.873$ & $\textbf{-0.304}$ & $68.105$ & $78.017$ & $87.383$ \\
& $\pm0.002$ & $\pm0.149$ & $\pm1.127$ & $\pm1.361$ & $\pm1.596$ \\
DAD-CSSE & $0.88$ & $-18.301$ & $60.02$ & $\textbf{68.181}$ & $100.173$ \\
& $\pm0.009$ & $\pm5.988$ & $\pm1.553$ & $\pm1.631$ & $\pm2.294$ \\
DAD-CASE & $0.873$ & $-13.745$ & $57.618$ & $\textbf{65.984}$ & $\textbf{103.226}$ \\
& $\pm0.004$ & $\pm8.163$ & $\pm0.833$ & $\pm0.96$ & $\pm1.492$ \\
DAD-SMIN & $0.873$ & $-10.474$ & $\textbf{59.827}$ & $\textbf{68.62}$ & $\textbf{99.951}$ \\
& $\pm0.009$ & $\pm5.252$ & $\pm1.709$ & $\pm2.26$ & $\pm3.25$ \\
DAD-BARR & $0.841$ & $-9.122$ & $61.751$ & $73.539$ & $93.535$ \\
& $\pm0.014$ & $\pm16.786$ & $\pm2.275$ & $\pm2.873$ & $\pm3.426$ \\
DAD-ALM & $0.875$ & $-12.002$ & $\textbf{57.403}$ & $\textbf{65.693}$ & $\textbf{103.74}$ \\
& $\pm0.009$ & $\pm4.775$ & $\pm0.737$ & $\pm1.083$ & $\pm1.737$ \\
\bottomrule
\end{tabular}
}
\label{proprietary_data_performance_reach}
\end{table}

\begin{table}[ht]
\centering
\caption{Results. Dataset II. K=5. In $AUROC_{\overline{y}}$ (higher is better), five proposed \texttt{DAD} models perform identical to two baselines \texttt{DISC}s. In $Within\ \%\ Budget$ (closer to 0 is better), \texttt{DAD-ALM} performs same as baseline \texttt{DISC-BC}. In $Spend\ per\ unit\ Reach$ (lower is better), \texttt{DAD-ALM} outperforms \texttt{DISC} baselines. In $E{\f}{\f}ective\ Spend$ (lower is better), \texttt{DAD-SMIN}, \texttt{DAD-ALM} outperform \texttt{DISC} baselines. In $Reach\ E{\f}{\f}c$-$E{\f}{\f}e$ (higher is better), \texttt{DAD-SMIN}, \texttt{DAD-ALM} outperform \texttt{DISC-BC} baseline. Overall, the \texttt{DAD} models show strong and significantly improved performance (shown in bold) over baselines \texttt{DISC}.  See Section~\ref{sec:reach_data2} for discussion.}
\label{google_performance_reach}
\resizebox{\columnwidth}{!}{%
\begin{tabular}{l|c|c|c|c|c}
\toprule
Model & AUROC  & Within & Spend per & Effective & Reach \\
    {} & $\overline{y}$ & \% Budget & unit Reach & Spend & Effc-Effe \\
\midrule
DISC-UC & $0.956$ & -- & $61.543$ & $64.333$ & -- \\
& $\pm0.002$ & -- & $\pm1.803$ & $\pm1.84$ & -- \\
DISC-BC & $0.956$ & $\textbf{-6.43}$ & $62.736$ & $65.594$ & $1640.449$ \\
& $\pm0.002$ & $\pm2.246$ & $\pm0.824$ & $\pm0.898$ & $\pm22.355$ \\
DAD-CSSE & $0.961$ & $-24.498$ & $63.471$ & $66.038$ & $1643.048$ \\
& $\pm0.001$ & $\pm6.616$ & $\pm2.189$ & $\pm2.304$ & $\pm58.492$ \\
DAD-CASE & $0.963$ & $-25.344$ & $62.285$ & $64.7$ & $1671.964$ \\
& $\pm0.001$ & $\pm9.345$ & $\pm1.826$ & $\pm1.919$ & $\pm48.866$ \\
DAD-SMIN & $0.957$ & $5.237$ & $59.229$ & $\textbf{61.884}$ & $\textbf{1745.79}$ \\
& $\pm0.002$ & $\pm4.846$ & $\pm1.629$ & $\pm1.707$ & $\pm44.765$ \\
DAD-BARR & $0.945$ & $-27.171$ & $62.667$ & $66.369$ & $1623.459$ \\
& $\pm0.006$ & $\pm4.207$ & $\pm1.009$ & $\pm1.309$ & $\pm30.659$ \\
DAD-ALM & $0.957$ & $\textbf{-3.485}$ & $\textbf{58.548}$ & $\textbf{61.196}$ & $\textbf{1757.545}$ \\
& $\pm0.001$ & $\pm3.955$ & $\pm0.749$ & $\pm0.738$ & $\pm21.582$ \\
\bottomrule
\end{tabular}
}
\end{table}

\begin{table}[]
\centering
\caption{Sensitivity to K. Dataset II. See Section~\ref{sec:google_sensitivity}. Similar to K=5 results, \texttt{DAD} models, namely \texttt{DAD-BARR} and \texttt{DAD-ALM} stand out in strong performance in 3 metrics as compared to baselines \texttt{DISC}s.} 
\label{google_K79_performance_reach}
\resizebox{\columnwidth}{!}{%
\begin{tabular}{l|c|c|c|c|c}
\toprule
Model & AUROC  & Within & Spend per & Effective & Reach \\
    {} & $\overline{y}$ & \% Budget & unit Reach & Spend & Effc-Effe \\
\midrule
K=7 & & & & & \\
\hline
DISC-UC & $0.955$ & -- & $60.605$ & $63.437$ & -- \\
& $\pm0.002$ & -- & $\pm2.26$ & $\pm2.358$ & -- \\
DISC-BC & $0.955$ & $\textbf{-3.787}$ & $63.113$ & $66.068$ & $1627.736$ \\
& $\pm0.002$ & $\pm1.55$ & $\pm0.633$ & $\pm0.701$ & $\pm17.448$ \\
DAD-CSSE & $0.961$ & $-19.716$ & $62.695$ & $65.234$ & $1661.574$ \\
& $\pm0$ & $\pm7.19$ & $\pm2.089$ & $\pm2.163$ & $\pm55.77$ \\
DAD-CASE & $0.962$ & $-24.853$ & $61.416$ & $63.842$ & $1698.022$ \\
& $\pm0.001$ & $\pm3.967$ & $\pm2.066$ & $\pm2.167$ & $\pm56.525$ \\
DAD-SMIN & $0.959$ & $6.298$ & $61.386$ & $63.984$ & $1691.268$ \\
& $\pm0.001$ & $\pm5.006$ & $\pm1.949$ & $\pm2.009$ & $\pm49.046$ \\
DAD-BARR & $0.947$ & $-43.356$ & $\textbf{60.354}$ & $\textbf{63.794}$ & $\textbf{1693.691}$ \\
& $\pm0.004$ & $\pm3.921$ & $\pm1.486$ & $\pm1.781$ & $\pm43.628$ \\
DAD-ALM & $0.962$ & $-13.002$ & $\textbf{57.89}$ & $\textbf{60.207}$ & $\textbf{1787.501}$ \\
& $\pm0.001$ & $\pm3.842$ & $\pm0.846$ & $\pm0.901$ & $\pm26.752$ \\
\hline
K=9 & & & & & \\
\hline
DISC-UC & $0.959$ & -- & $59.482$ & $61.99$ & -- \\
& $\pm0.001$ & -- & $\pm1.596$ & $\pm1.645$ & -- \\
DISC-BC & $0.959$ & $\textbf{-4.024}$ & $63.656$ & $66.346$ & $1621.532$ \\
& $\pm0.001$ & $\pm1.629$ & $\pm0.772$ & $\pm0.837$ & $\pm20.866$ \\
DAD-CSSE & $0.962$ & $-29.568$ & $58.923$ & $\textbf{61.272}$ & $\textbf{1757.175}$ \\
& $\pm0.001$ & $\pm4.949$ & $\pm0.958$ & $\pm1.013$ & $\pm29.456$ \\
DAD-CASE & $0.963$ & $-28.462$ & $59.628$ & $\textbf{61.897}$ & $\textbf{1744.099}$ \\
& $\pm0.001$ & $\pm6.72$ & $\pm1.563$ & $\pm1.592$ & $\pm41.464$ \\
DAD-SMIN & $0.957$ & $1.845$ & $62.286$ & $65.079$ & $1671.734$ \\
& $\pm0.002$ & $\pm6.86$ & $\pm2.438$ & $\pm2.582$ & $\pm66.204$ \\
DAD-BARR & $0.955$ & $-40.699$ & $\textbf{60.307}$ & $\textbf{63.168}$ & $\textbf{1703.421}$ \\
& $\pm0.001$ & $\pm5.292$ & $\pm0.846$ & $\pm0.915$ & $\pm23.898$ \\
DAD-ALM & $0.962$ & $\textbf{-3.617}$ & $\textbf{58.102}$ & $\textbf{60.405}$ & $\textbf{1781.633}$ \\
& $\pm0.001$ & $\pm4.486$ & $\pm0.887$ & $\pm0.908$ & $\pm26.626$ \\
\bottomrule
\end{tabular}
}
\end{table}

\subsection{Results, Dataset II, Public}
\label{sec:reach_data2}
Throughout all tables, results are shown with \textbf{Mean +/- Stderr} based on 8 seeds. Caption of Table~\ref{google_performance_reach} highlights the significantly improved performance of \texttt{DAD} models over SOTA baselines, DISCs. Note that two metrics are undefined for DISC-UC. Overall, DAD-ALM with strong performance in four metrics stand out. Similar to Dataset I, results with Public, Dataset II
strongly reinforce that decreased spend for maximizing reach is achievable with no depreciation in performance in conversion prediction. That is, delivery aware discovery outweighs traditional discovery which ignore delivery and budget constraint. We now move on to sensitivity experiments by varying $K$.

\subsection{Sensitivity to $K$} \label{sec:google_sensitivity}
Sensitivity to the choice of $K$, number of segments, with Dataset II, public data, is shown in Table~\ref{google_K79_performance_reach}, for K=7 and 9. \texttt{DAD-BARR} and \texttt{DAD-ALM} strongly outperform baselines in crucial composite metrics \textit{Effective Spend} and \textit{Reach-Effc-Effe}, and perform the same in predicting conversion, $AUROC_{\overline{y}}$. Comparing with Table~\ref{google_performance_reach} finds remarkable consistency across K=5, 7, 9, strongly favoring our proposed models over SOTA baselines.   


\subsection{Ablation Study}
\label{sec:ablation_both_data}

The ablation study checks whether the proposed full model, or, the full network architecture shown in Figure~\ref{fig:overview} is necessary to achieve our dual goals of strong predictive performance in conversion and reach maximization, subject to budget constraint. Or, can a reduced architecture give comparable performance? To test this, we \textit{turn off} Step 2 of the architecture in training, that is, did not train Actor(Encoder, Selector)-Critic(Predictor) and centroid Embeddings (see green text in Figure~\ref{fig:overview}). We trained Steps 1 and 3 only. The ablated results, with Dataset II, are shown in Table~\ref{google_ablation_K579_sensitivity} for each $K$ = 5, 7, 9. We compare the full architecture's performance with that of the reduced architecture, for each $K$, affording generalization of the ablation. As the caption in Table~\ref{google_ablation_K579_sensitivity} states, the ablated results show large decrease in performance compared to the full architecture, for each $K$ = 5, 7, 9. This strongly justifies the use of the proposed full architecture for the delivery aware discovery problem.

\begin{table}[]
\centering
\caption{Ablation Sensitivity to K. Dataset II. See Section~\ref{sec:ablation_both_data}. For each K, comparing results of the full model (see Tables~\ref{google_performance_reach} and~\ref{google_K79_performance_reach}), the ablated results below show that: Each \texttt{DAD} model has appreciably large (i) decrease in $AUROC_{\overline{y}}$, (ii) increase in $E{\f}{\f}ective\ Spend$, and (iii) decrease in $Reach\ E{\f}{\f}c$-$E{\f}{\f}e$.} 
\label{google_ablation_K579_sensitivity}
\resizebox{\columnwidth}{!}{%
\begin{tabular}{l|c|c|c|c|c}
\toprule
Model & AUROC  & Within & Spend per & Effective & Reach \\
    {} & $\overline{y}$ & \% Budget & unit Reach & Spend & Effc-Effe \\
\midrule
K=5 & & & & & \\
\hline
DAD-CSSE & 0.898    & -31.965         & 61.615          & 68.606          & 1575.407          \\
         & $\pm0.004$ & $\pm3.561$        & $\pm1.672$        & $\pm1.945$        & $\pm42.224$         \\
DAD-CASE & 0.898    & -28.608         & 62.268          & 69.374 & 1561.016 \\
         & $\pm0.004$ & $\pm5.573$        & $\pm1.984$        & $\pm2.196$        & $\pm49.403$         \\
DAD-SMIN & 0.866    & \textbf{2.948}  & 60.238 & 69.716 & 1546.403          \\
         & $\pm0.015$ & $\pm1.597$        & $\pm1.126$        & $\pm1.479$        & $\pm32.301$         \\
DAD-BARR & 0.857    & -41.1  & \textbf{58.436} & 68.221 & 1577.539          \\
         &  $\pm0.008$ & $\pm4.475$        & $\pm0.794$        & $\pm1.01$         & $\pm23.999$         \\
DAD-ALM  & 0.898    & -11.66 & \textbf{58.997} & \textbf{65.7}   & \textbf{1641.753} \\
         & $\pm0.001$ & $\pm2.53$         & $\pm1.352$        & $\pm1.546$        & $\pm35.428$  \\
\hline
K=7 & & & & & \\
\hline
DAD-CSSE & $0.9$ & $-26.207$ & $59.284$ & $65.847$ & $1637.582$ \\
         & $\pm0.003$ & $\pm4.598$ & $\pm1.357$ & $\pm1.429$ & $\pm35.284$ \\
DAD-CASE & $0.9$ & $-15.201$ & $61.319$ & $\textbf{68.114}$ & $\textbf{1582.834}$ \\
         & $\pm0.003$ & $\pm4.831$ & $\pm1.258$ & $\pm1.425$ & $\pm33.951$ \\
DAD-SMIN & $0.877$ & $\textbf{-2.069}$ & $\textbf{61.442}$ & $\textbf{70.08}$ & $1539.038$ \\
         & $\pm0.005$ & $\pm2.513$ & $\pm1.479$ & $\pm1.571$ & $\pm34.241$ \\
DAD-BARR & $0.876$ & $\textbf{-36.016}$ & $\textbf{62.338}$ & $\textbf{71.203}$ & $1520.433$ \\
         & $\pm0.005$ & $\pm4.633$ & $\pm1.776$ & $\pm2.261$ & $\pm46.257$ \\
DAD-ALM  & $0.891$ & $\textbf{-18.332}$ & $\textbf{58.468}$ & $\textbf{65.602}$ & $\textbf{1639.33}$ \\
         & $\pm0.001$ & $\pm5.067$ & $\pm0.69$ & $\pm0.758$ & $\pm19.304$ \\
\hline
K=9 & & & & & \\
\hline
DAD-CSSE & $0.896$ & $-27.864$ & $58.868$ & $65.673$ & $1637.147$ \\
& $\pm0.003$ & $\pm2.485$ & $\pm0.571$ & $\pm0.658$ & $\pm16.876$ \\
DAD-CASE & $0.894$ & $-26.952$ & $59.551$ & $66.605$ & $1620.766$ \\
& $\pm0.003$ & $\pm6.121$ & $\pm1.517$ & $\pm1.725$ & $\pm38.121$ \\
DAD-SMIN & $0.819$ & $\textbf{11.684}$ & $\textbf{62.404}$ & $\textbf{76.973}$ & $1416.837$ \\
& $\pm0.03$ & $\pm6.11$ & $\pm1.947$ & $\pm3.41$ & $\pm60.135$ \\
DAD-BARR & $0.859$ & $\textbf{-36.784}$ & $\textbf{61.994}$ & $\textbf{72.136}$ & $1498.258$ \\
& $\pm0.003$ & $\pm3.793$ & $\pm1.736$ & $\pm2.005$ & $\pm40.722$ \\
DAD-ALM & $0.896$ & $\textbf{-12.456}$ & $\textbf{58.753}$ & $\textbf{65.592}$ & $\textbf{1639.492}$ \\
& $\pm0.003$ & $\pm3.386$ & $\pm0.74$ & $\pm0.74$ & $\pm18.771$ \\
\bottomrule
\end{tabular}
}
\end{table}

\subsection{Evaluation of Beh2Stat} \label{sec:beh2stat_test}

\begin{table}[h!]
\centering
\caption{Evaluation of Beh2Stat function. See section~\ref{sec:beh2stat_test}.
}
\label{table:intermediate-evaluation-of-beh2stat}
\begin{tabular}{@{}r c|cc}
\toprule
&&
\multicolumn{2}{c}{\textsc{Accuracy}}
\\
\cmidrule{3-4}
\multicolumn{1}{r}{\textbf{Static Characteristics}}   &
\multicolumn{1}{c}{\textbf{\# Classes}} &
\multicolumn{1}{c}{\textbf{Beh2Stat}} &
\multicolumn{1}{c}{\textbf{Random}}
\\
\midrule
\textit{Country} & 6	& 0.58 &	0.17
\\
\textit{Source} & 2 &	0.86	& 0.5
\\
\textit{Member} & 3 &	0.53 &	0.33
\\
\textit{Browser} & 6	& 0.51 &	0.17
\\
\textit{Operating System (OS)} & 6 &	0.51	& 0.17
\\
\bottomrule
\end{tabular}
\end{table}

For the proprietary data, Table~\ref{table:intermediate-evaluation-of-beh2stat} shows that Beh2Stat has accuracies appreciably higher than that of the random model. Random prediction accuracy values are shown in the last column, which vary by the number of classes.

\section{Conclusion}
\label{sec:conc}
Behavioral segmentation of users is highly desirable by a firm as emphasized in both industry and academia ~\cite{hubspot_beh_seg, gupta2014marketing} since users' online behaviors are very predictive of outcomes such as conversion. Firms send different offers, messages, communications to different segments.
The obstacle in delivering messages on media channels lies in reaching these \textit{behavioral} segments on media.
On a medium, only a proportion of a behavior segment can be matched, and of those matched only a fraction sees / clicks on a message.
Moreover, on a medium, users are defined by their static characteristics, but not by their online interactions with the firm, which only the firm knows, but the media do not.
For discovery to be useful, delivery ought to be considered simultaneously, and not sequentially after discovery.
Extant work on discovery of user segments ignore the need for this simultaneity between discovery and delivery.
We offer an approach to fill this research gap. Extensive experiments on two datasets - proprietary and public (Google Analytics) - find strong support for our approach. Moreover, sensitivity experiments on Google data, by varying the hyperparameter, number of segments, affirm those findings that our approach achieves high, improved performance on Spend and Reach metrics, while achieving equally good predictive conversion performance AUROC, compared to two SOTA baselines.  
Ablation studies, including sensitivity of ablation to number of segments, further justify the value of our proposed network to address this joint delivery-aware-discovery optimization problem at hand. 

As a way of limitation, our joint optimization works when the cost of a message sent to a user is known and not an outcome of bidding. Future research may address this problem, under ad bidding based media spend and budget pacing.

\balance
\bibliographystyle{ACM-Reference-Format}
\bibliography{refs}



\end{document}